# FANET Experiment: Real-Time Surveillance Applications Connected to Image Processing System

**Bashir Olaniyi Sadiq[1,2], Muhammed Yusuf Abiodun[2], Sikiru Olayinka Zakariyya[3], and Mohammed Dahiru Buhari[2,4]**

[1] *Computer Engineering Department, Ahmadu Bello University, Nigeria*
[2] *Electrical, Telecommunication and Computer Engineering Department, KIU, Uganda;*
[3] *Electrical and Electronic Engineering Department, University of Ilorin, Kwara State, Nigeria*
[4] *Electrical and Electronic Engineering Department, Abubakar Tafawa Balewa University, Bauchi State, Nigeria*

**\*** *Correspondence:* bosadiq@kiu.ac.ug

**Abstract**
*The major goal of this paper is to use image enhancement techniques for enhancing and extracting data in FANET applications to improve the efficiency of surveillance. The proposed conceptual system design can improve the likelihood of FANET operations in oil pipeline surveillance, and sports and media coverage with the ultimate goal of providing efficient services to those who are interested. The system architecture model is based on current scientific principles and developing technologies. A FANET, which is capable of gathering image data from video-enabled drones, and an image processing system that permits data collection and analysis are the two primary components of the system. Based on the image processing technique, a proof of concept for efficient data extraction and enhancement in FANET situations and possible services is illustrated.*

**Nomenclature and units**

| | |
|---|---|
| RGB | Red, Green and Blue |
| FANET | Flying Ad Hoc Network |
| PRoFFAN | Priority-based Routing Framework for Flying Adhoc Networks |
| UAV | Unmanned Aerial Vehicle |
| BS | Base Station |
| VANET | Vehicular Ad hoc network |
| CUV | Collaborative scheme between UAVs and VANETs |
| SIFT | Scale-Invariant Feature Transform |
| GUI | Graphical User Interface |

## 1.0 Introduction

Environmental research can be more reliable with real-time surveillance. The idea is to keep an eye on a specific area to gather information about the target area. To put it another way, the real-time surveillance challenge might be reframed as an aerial surveillance or area coverage problem. Static monitoring instruments, such as cameras, are employed to cover the entire region in a classic area coverage problem (Sadiq *et al*., 2020; Felemban *et al*., 2021). The situation is different when it comes to environmental research issues. This is because the job area in the environmental research scenario is so large that the static sensors may be unable to reach the intended location. In this application, mobility vehicles with sensors, particularly aerial vehicles with cameras, are favored over static sensors. The use of more than one aerial vehicle with cameras cooperating to achieve tasks is known as Flying Ad hoc Network (FANET) (Fang *et al*., 2020). FANET drones have a high degree of mobility and maneuverability, allowing them to go where humans cannot. It is frequently tasked with jobs that are regarded as filthy or harmful by the general public. Drones were created for military purposes; hence their size is fairly large, however, after many years of evolution, their size is shrinking. On the civil market, more drones with sensors, such as cameras, sonars, and lidars, are being introduced. Drones have proven to be quite effective in environmental studies, such as gathering information about a certain area to learn about the species (Ilker *et al*., 2015; Bilal *et al*., 2016; Zhiqing et al., 2019). Because of their low cost, great mobility, and real-time data transmission capabilities, these drones are regarded as the best choice for environmental investigations (Sahingoz, 2014; Sadiq *et al*., 2018; Zhang, and Li, 2020). Figure 1 shows the FANET drone applications in different scenarios.

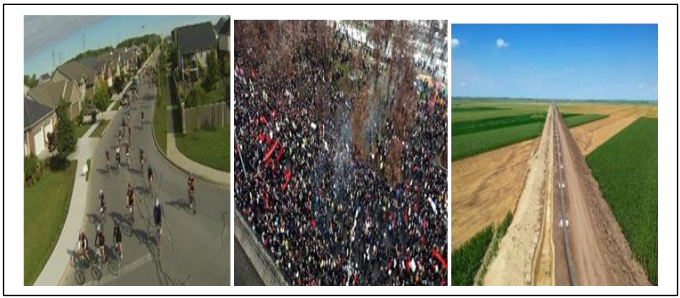

**Figure 1** FANET applications scenario (**a**) A Sports Coverage, (**b**) A Media Coverage, (**c**) An Oil Pipeline Surveillance *Felemban et al., 2021*.

Real-time photographs and movies recorded by drones will have a wider field of view, cover large areas, have a minimal blockage in open places, and have an adjustable alleviating view, as shown in the application scenarios in Figure 1 when operated in an ad hoc manner (Sadiq *et al*., 2018). The photos collected by the UAVs are sent to the end system using standard FANET communication media. Before the FANET programs can process the image, they must first get the whole image. Image-processing applications necessitate the transfer of huge volumes of data between the UAV and the end system. To enable communication between two UAVs and the transfer of image data, some works have suggested the use of antennas (Zakariyya *et al*., 2016; Sadiq and Salawudeen, 2020; Jamlos *et al*., 2020). The image data acquired by FANET for enhancement or extraction is measured in kilobytes or even megabytes. A raw Red Green–Blue (RGB) image with 24-bits per pixel (8 bits per color) of 128x128 pixels will be 128x128x24 = 393,216 bits (approximately 48 kilobytes) (Olaniyi *et al*., 2018). The FANET drone chosen for this research work is the Tello-Edu Drone. Tello EDU is a programmable drone designed for educational use. The Tello EDU was designed by Shenzhen Ryze Technology and includes DJI flight control technology and Intel processors. Tello existed before this version, but unlike this new architecture, it did not allow for simple swarm programming. In addition, this edition has a new SDK that is more widely available than the previous one. The diagram of the used Tello Edu drone for this research work is depicted in Figure 2.

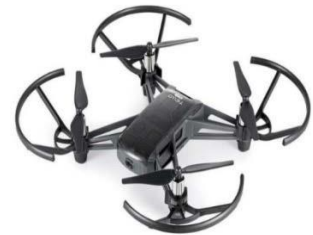

**Figure 2** Used Tello-Edu Drone

This research offers a FANET-based image improvement and extraction method for Flying Area Networks that will improve the quality of important image features from a UAV to the end system. Images from oil pipeline monitoring, sport, and media coverage are our motivating applications, and we hope that getting essential picture attributes will improve the use of FANET drones in these application domains. Using the programmable Tello-Edu drone, we validated the transmission from UAVs to the end system in this study. In MATLAB, we implemented our conceptual framework.

A Priority-based Routing Framework for Flying Adhoc Networks (PRoFFAN) is presented in Felemban *et al.,* (2021) for the rapid transmission of critical multimedia data to control centers. By prioritizing the transmitting and forwarding of essential image data from the UAV to the control center, PRoFFAN speeds up the FANET application's reaction time. Crowd management is the authors' motivational application; we believe that providing crucial image features as soon as possible will save lives and improve crowd safety and flow. nonetheless, there is a need to enhance the images in real-time to improve viewed crowed. The work of Rabahi *et al*., (2020) presented UAVs- Based Mobile Radars for Real-Time Highways Surveillance. The presented work planned to utilize radar drones to monitor highways for traffic offenses and then send the data to a ground police patrol or a police data center. The acquired images are relayed directly to the ground Base Station (BS) if the UAV fleet is close enough. If the fleet is far away, however, the UAVs must return to the BS, which is a major worry for UAVs because their batteries have a finite lifespan. The real- time feature of this application is ensured by the collaboration of UAVs with ground vehicles. The findings of the presented work which was based on simulation only reveal that teamwork produces better results than when simply UAVs are used. In Al Fayez *et al*., (2019), the technological capabilities of existing and emerging surveillance systems utilized for international border monitoring applications are investigated. It discusses the efficacy of these systems as well as the technological infrastructure needed to deploy them. Identification of these systems' strengths and limitations, as well as their ability to tackle present and future challenges, has received special attention. It was justified by the authors that the reliability of a large-scale border surveillance system in terms of monitoring coverage, activity detection accuracy, and real-time operation is critical to its effectiveness. Such objectives must be satisfied at cheap hardware and operation costs, in difficult weather and terrain conditions, with minimal human participation, and in the face of frequent hardware failures. Due to its dynamical sparse linear architecture, a fleet of drones en route to highway monitoring may experience frequent link failures. A collaboration of UAVs with VANETs would be a suitable solution to compensate for the sparse topology, however, integration of the two networks could cause too much overhead. As such, the authors in the work of Bashir and Boudjit, (2020) proposed a proactive energy-efficient, and reliable Collaborative scheme between UAVs and VANETs termed CUV. The suggested CUV protocol is an innovative way of addressing difficulties unique to UAV networks. The CUV is a collaborative strategy for UAVs that seeks assistance from VANETs when local UAVs are unavailable. The findings showed that CUV is an energy-efficient and dependable approach for dealing with. The authors claim in Michelett *et al*, (2018) that flying ad-hoc networks can provide the communication assistance required in disaster relief settings, and the authors present a new approach to accomplishing this goal. The concept calls for the employment of flying ad-hoc units, which were executed using drones and would serve as communication hubs for first responders working in different parts of the impacted area. The Flying Real-Time Network is the name of the proposed technique, and its feasibility in providing communication in a disaster scenario is demonstrated by presenting both a real-time schedulable analysis of message delivery and simulations of communication support in a physical scenario based on a real incident. The UAV image mosaic for the road traffic accident scene was presented by the authors in the work of Liu *et al*., (2019). The authors broke down the focus of UAV image mosaic technology into two sections: image registration and picture fusion. The authors demonstrated intelligent distance measurement in a road traffic collision scene using the UAV image mosaic technique. Mean Filter, Poisson, or Laplacian distribution methods were used to remove noise. The images in the UAV aerial pictures mosaic have a high resolution and a big number of them. The authors employed the SIFT method in this paper. Despite its resistance to noise, angle of view, and light, the SIFT technique requires a considerable amount of computation, has a high temporal complexity, and is inefficient. However, the images were not linked to the UAV and were not processed in real-time. By combining the advantages of unmanned aerial vehicle (UAV) remote sensing data and image recognition technology, the authors in the work of Gao *et al*., (2019) suggested a novel measuring technique. To begin, a UAV airborne camera acquired footage of the water fluctuation process, and image recognition technology was used to distinguish and extract the water surface line in the imagery. After then, parameter calibration was used to compute successive water levels at a measuring segment. During the capturing time, statistical metrics of water levels such as maximum, average, and minimum values were calculated. The authors integrated UAV photogrammetry and image recognition technology developed a new optical measuring technology to measure water level in the field, and built the corresponding measurement system, which included UAV, airborne camera, plane wall, baffle or straight line, calibration points, correction point, and image processing software, to overcome the limitations of current measuring instruments and methods for water level measurements in the field.

Based on the reviewed work, it is evident that limited or no works of literature have directly tried to embark on the real-time processing of the captured images when employed in civilian applications such as sports and media coverage to enhance and improve visualization. As such, this paper presents an experimental study of real-time image enhancement of drone-captured images in real-time to further strengthen the UAV applications scenario

## 2.0    Methodology

The technique of capturing a picture at the camera node is required by some FANET applications. FANET drones capture images with a camera placed on a UAV and transfer them to the end system. Through relay nodes that act as intermediaries, information can be transmitted from the camera node to the end system in sequential order. Traditionally, packets are not prioritized based on their kind, therefore crucial information may be delayed in reaching the end system. We present a mechanism for transmitting image data in FANET using image processing systems in this study. The methodology used in this paper is shown in Figure 3. UAVs capture photos for FANET, which must then be improved and communicated. For enhancement, the acquired image is sent to the image processing toolbox.

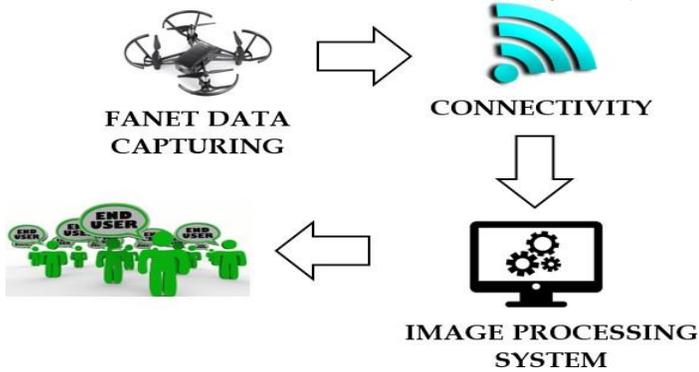

**Figure 3** FANET Methodology Used

Before deploying the FANET drones (hence referred to as "Tello-Edu Drones"), the first step is to wirelessly connect the drones to the image processing system, which in this case is a computer system. The link was made using python shell scripts that opened the drone's and system's UDP ports for communication. The drones were commanded autonomously for the mission they were deployed for using the same python shell script. All of the images produced or collected by the drones can be viewed in real-time via the associated computer system once communication is achieved. To enhance the images collected from the drones, a standalone image processing application was developed using the MATLAB toolbox as depicted in Figure 4.

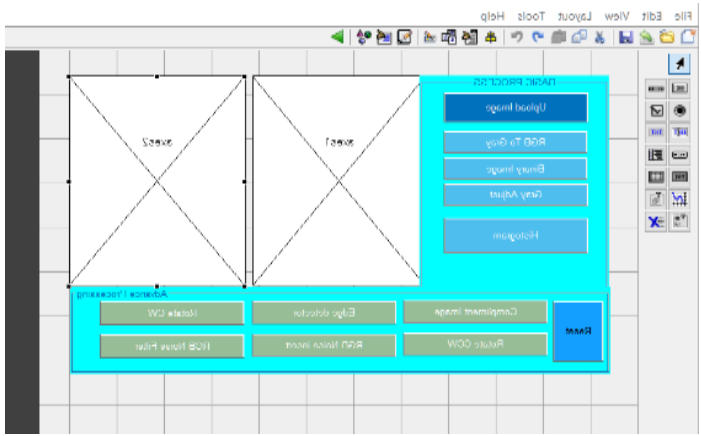

**Figure 4** User Interface of the Proposed system

The drone capture images were easily accessed and integrated into the created GUI image processing system for image alteration and enhancement. Colored to grayscale conversion, picture compliment, noise filter, gray adjust, histogram, edge detection, and image rotation are some of the capabilities offered in the built GUI-based image processing application.

The FANET experimentation in this work focuses on single-hop picture transmission via the image processing system between the UAV (Camera Node) and the end station. As a result, there is only one UAV in range of the end station. Within a limited area, each UAV travels a distance of 10 meters. The implementation algorithm used for the methodology is presented in algorithm 1

**Algorithm 1** Implementation algorithm

1:     *Open drone connection port* = 8889
2:     **for** *i ≤ n* **do**
3:             **if** *i* == 1 **then**
4:                     *connect*
5:                     **end if**
6:     **end for**
7:     send flight command
8:     send captured images to the system
9:     processing images
10:   stop

## 3.0    Experimentation

Python idle and MATLAB are the two main pieces of software used in this experiment. The main controller program used to interface the Tello drone with certain dependencies is Python Idle. MATLAB is used to set up the GUI to handle the image processing phase of the work. The connection between the system is achieved via the diagram presented in Figure 5.

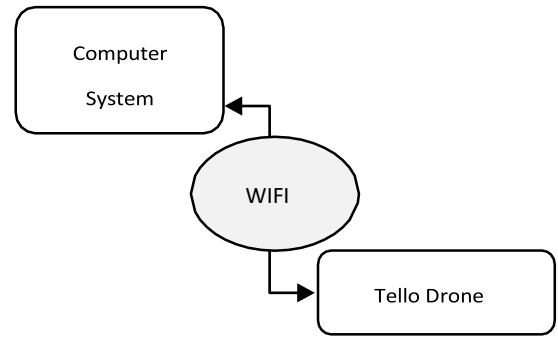

**Figure 5** Connection Diagram

Various operations were carried out to meet the system's core goal, which is the fundamental Tello mission, as well as Tello video and image acquisition (image processing). The initial step, as illustrated in Figure 5, is to link the drone to the computer so that a python script may be used to issue commands. The connection between the drone and the PC must be established via Wi-Fi. Then instructions are given to take off, complete a task (in this case, a square), and land. The socket library was imported and was used to build the various connections. After that, the authors define the local address (IP: 0.0.0.0 Port: 9000) and then construct and bind the socket. The next step is to determine the Tello EDU drone's address: IP: 192.168.10.1 Port: 8889, and we can send commands to the address using the socket that we generated earlier. To transmit the various commands, we'll utilize the python command socket. sendto'. The Tello command and the tello address will be the arguments. When the drone camera is turned on and images are taken while the drone is on a mission, this is where the indoor inspection and media coverage operations come into action. The collected image was accessible and imported into a graphical user interface in MATLAB for image modification and processing in this phase of the study. RGB2Gray, picture compliment, noise filter, gray adjust, histogram, edge detection, and rotation are the sample of the image processing steps used. While on the mission, Tello video streams the real-time image, and a snap button is offered to take a snapshot if necessary. The sample snapshot image is shown in Figure 6.

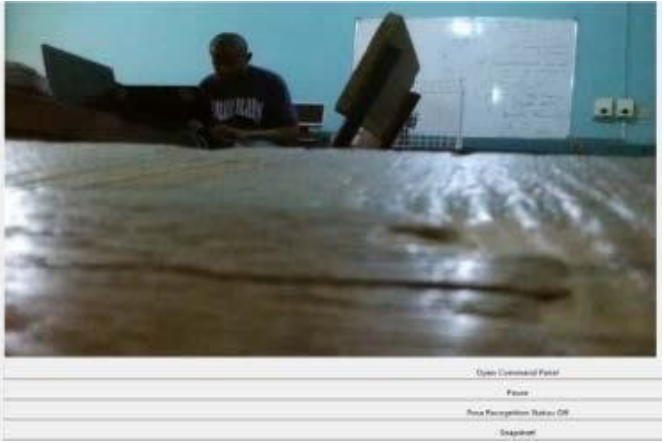

**Figure 6** The sample Snapshot of the Captured image

The outcomes of running some of the snapshot im- ages through the design graphical user interface (Image processing system) shown in Figure 7 are summarized hereunder.

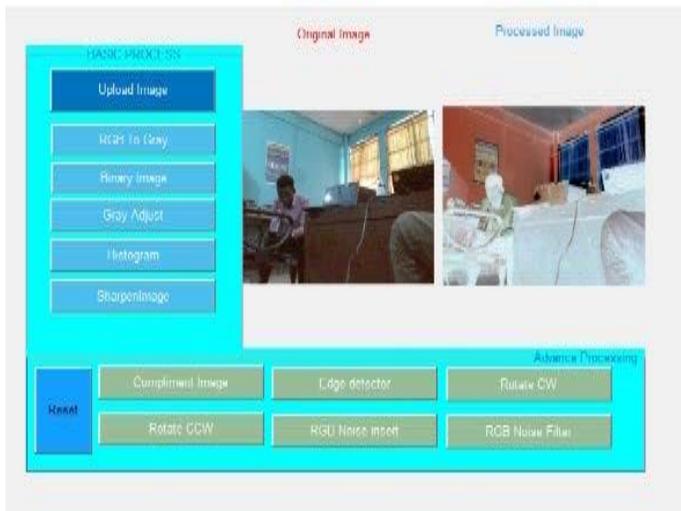

**Figure 7** GUI of the Image Processing System

By removing the hue and saturation information while keeping the luminance, the rgb2gray function converts RGB images to grayscale. The RGB image has been transformed to grayscale in the real-time experiment presented in Figure 8(a). In a binary image's complement, zeros become ones and ones become zeros. The colors black and white have been inverted. Each pixel value in a grayscale or color image is deducted from the maximum pixel value in the complement. The image that has been complimented is shown in Figure 8(b). Only binary images are suitable for edge detection. By evaluating only, the edge elements rather than all pixels, the processing time is minimized. The most common applications of edge detection are feature extraction and detection. To detect the presence of an edge in an image, the magnitude of the first derivative is calculated within a neighbor- hood. the pixel of interest is used. The edge is detected using gradient operators such as Sobel, Prewitt, and Canny. The sample image on edge detection is shown in Figure 8(c). The incidence or frequency of a given gray level is represented by the histogram of a picture. It's a graph with gray level intensities on the x-axis and their frequency on the y-axis. It reveals information about an image's contrast. It aids in the classification of photographs. Image statistics are provided for several approaches such as thresholding, intensity slicing, and segmentation. It's useful to know if the digitizer's complete dynamic range is being utilized. The image histogram result is shown in figure 8(d).

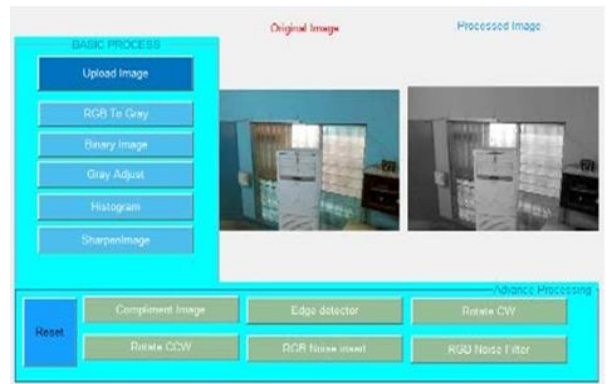

(a)

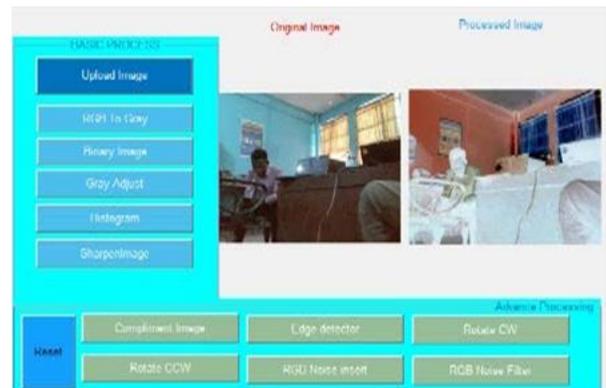

(b)

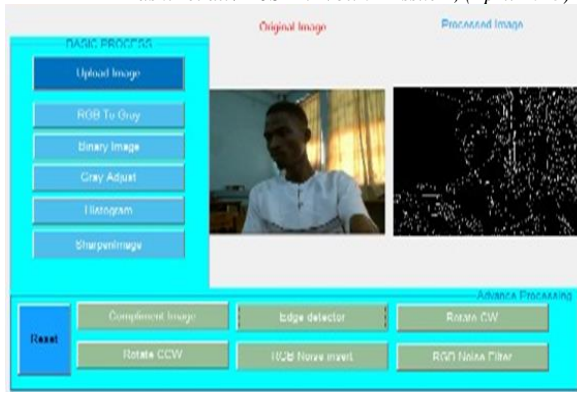

(c)

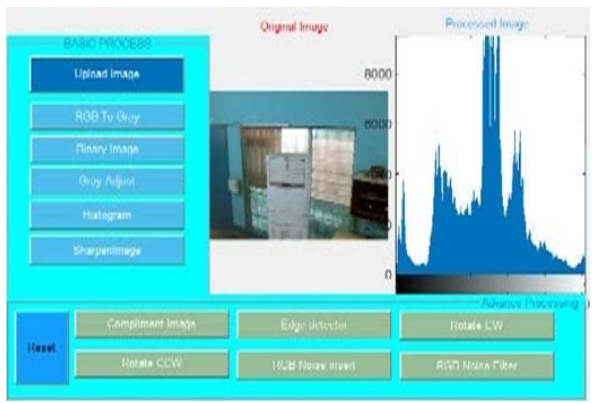

(d)

**Figure 8** Results of the Real-Time Experiment Using the Image Processing System

The Tello drone was used to take images in both remote and indoor locations. The Tello-Edu drone is operated by a computer system that uses a series of python scripts, and MATLAB was used to create a graphical user interface for processing images collected from the drone coverage Various image processing techniques were implemented, including RGB2Gray, Edge Detection, and Image Complement, among others. The Experiment shows that the FANET captured images can be processed in real-time which makes the application of FANET to indoor sport and media coverage efficient in the presence of varying inherent noise characteristics such as speckle noise, blurring, and dust amongst others

## 4.0    Conclusions

The research's main goal, which was to apply image enhancement algorithms to enhance and extract data in FANET applications to improve surveillance efficiency, was largely achieved. The research team developed a conceptual system architecture that has the potential to improve FANET operations in oil pipeline surveillance, sports, and media coverage, with the ultimate goal of offering efficient services to interested people. The system was constructed around two key components: a FANET capable of gathering detection data from video-enabled drones and an image processing system that allows data collection and analysis. A proof of concept for effective data extraction and augmentation in FANET settings, as well as possible services, was demonstrated using the image processing technique.

## Acknowledgements
Funding for the purchase of Drones was sponsored by the Institutional Based Research Grant, Ahmadu Bello University Zaria, Nigeria.

## Declaration of conflict of interest
I hereby declare that there is no known conflict of interest.

Bashir et al. / KJSET: Vol. 02 Issue 1, (April 2023) 10-16, ISSN: 1958-0641, https://doi.org/10.59568/KJSET-2023-2-1-02